\documentclass[11pt]{article}
\usepackage[a4paper]{geometry}
\usepackage{networds2015}
\usepackage{times}
\usepackage{url}
\usepackage[colorlinks=true, linkcolor=black, urlcolor=blue, citecolor=black, anchorcolor=black]{hyperref}
\usepackage{latexsym}
\usepackage{graphicx}
\usepackage{natbib}
\usepackage{amsmath,amssymb}
\usepackage{subfig}
\title{A Distributional Semantics Approach to Implicit Language Learning}

\author{Dimitrios Alikaniotis \qquad John N. Williams \\
  Department of Theoretical and Applied Linguistics \\
  University of Cambridge \\
  9 West Road, Cambridge CB3 9DP, United Kingdom \\
  {\tt \{da352|jnw12\}@cam.ac.uk} \\}

\date{}

\begin{document}
\maketitle

\begin{abstract}
  In the present paper we show that distributional information is
  particularly important when considering concept availability under
  implicit language learning conditions. Based on results from
  different behavioural experiments we argue that the implicit
  learnability of semantic regularities depends on the degree to which
  the relevant concept is reflected in language use. In our
  simulations, we train a Vector-Space model on either an English or a
  Chinese corpus and then feed the resulting representations to a
  feed-forward neural network. The task of the neural network was to
  find a mapping between the word representations and the novel
  words. Using datasets from four behavioural experiments, which used
  different semantic manipulations, we were able to obtain learning
  patterns very similar to those obtained by humans.
\end{abstract}

\section{Introduction}\label{sec:introduction}

Vector-space models of semantics (VSMs) derive word representations by keeping track of the co-occurrence patterns of each word when found in large linguistic corpora.
By exploiting the fact that similar words tend to appear in similar contexts \citep{Harris:1954kg}, such models have been very successful in tasks of semantic relatedness \citep{Landauer:1997tc,Rohde:2006tz}.
A common criticism addressed towards such models is that those co-occurrence patterns do not explicitly encode specific semantic features unlike more traditional models of semantic memory \citep{Collins:1969kr,Rogers:2004}.
Recently, however, corpus studies \citep{Bresnan:2008gs,Hill:2013cv} have shown that some `core' conceptual distinctions such as animacy and concreteness are reflected in the distributional patterns of words and can be captured by such models \citep{Hill:2013uv}.

In the present paper we argue that distributional characteristics of words are particularly important when considering concept availability under implicit language learning conditions.
Studies on implicit learning of form-meaning connections have highlighted that during the learning process a restricted set of conceptual distinctions are available such as those involving animacy and concreteness.
For example, in studies by \citet{Williams:2005cg} (W) and \citet{Leung:2014bi} (L\&W) the participants were introduced to four novel determiner-like words: \emph{gi}, \emph{ro}, \emph{ul}, and \emph{ne}.
They were explicitly told that they functioned like the article `\emph{the}' but that \emph{gi} and \emph{ro} were used with near objects and \emph{ro} and \emph{ne} with far objects.
What they were not told was that \emph{gi} and \emph{ul} were used with living things and \emph{ro} and \emph{ne} with non-living things.
Participants were exposed to grammatical determiner-noun combinations in a training task and afterwards given novel determiner-noun combinations to test for generalisation of the hidden regularity.
W and L\&W report such a generalisation effect even in participants who remained unaware of the relevance of animacy to article usage --– semantic implicit learning.
\citet{paciorek2015} (P\&W) report similar effects for a system in which novel verbs (rather than determiners) collocate with either abstract or concrete nouns.
However, certain semantic constraints on semantic implicit learning have been obtained.
In P\&W generalisation was weaker when tested with items that were of relatively low semantic similarity to the exemplars received in training.
In L\&W Chinese participants showed implicit generalisation of a system in which determiner usage was governed by whether the noun referred to a long or flat object (corresponding to the Chinese classifier system) whereas there was no such implicit generalisation in native English speakers.
Based on this evidence we argue that the implicit learnability of semantic regularities depends on the degree to which the relevant concept is reflected in language use.
By forming semantic representations of words based on their distributional characteristics we may be able to predict what would be learnable under implicit learning conditions.

\section{Simulation}\label{sec:simulation}

\begin{figure}[t]
	\centering
			\includegraphics[width=\columnwidth]{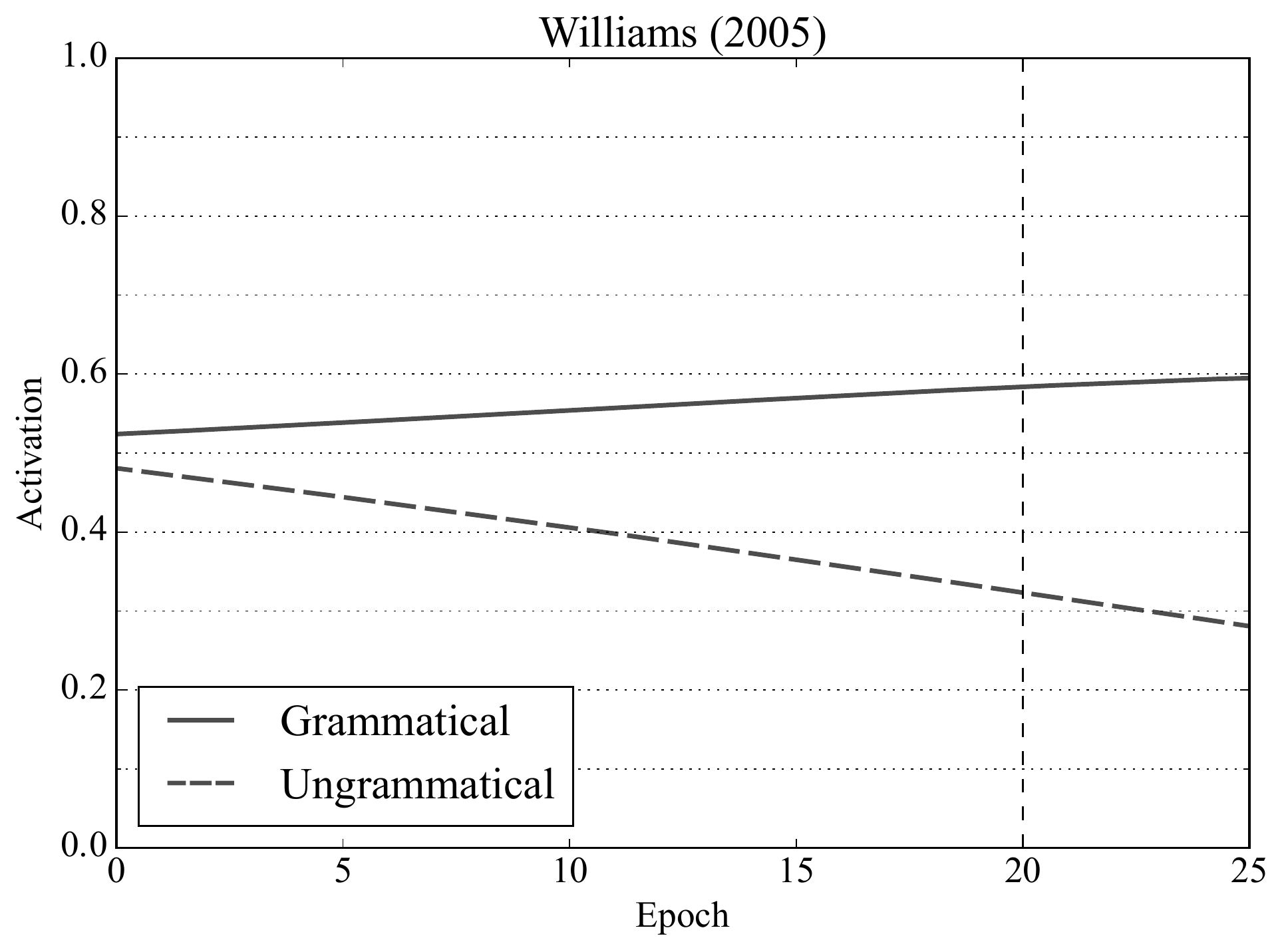}
			\caption{Generalisation gradients obtained from the Williams (2005) dataset. The gradients were obtained by averaging the output activations for the grammatical and the ungrammatical pairs, respectively. The network hyperparameters used were: learning rate: $\eta = 0.01$, weight decay: $\gamma = 0.01$, size of hidden layer: $\mathbf{h} \in \mathbb{R}^{100}$. For this and all the reported simulations the dashed vertical lines mark the epoch in which the training error approached zero. See text for more information on the experiment.}
\end{figure}

\begin{figure}[t]
	\centering
			\begin{minipage}{\columnwidth}
        \includegraphics[width=\columnwidth]{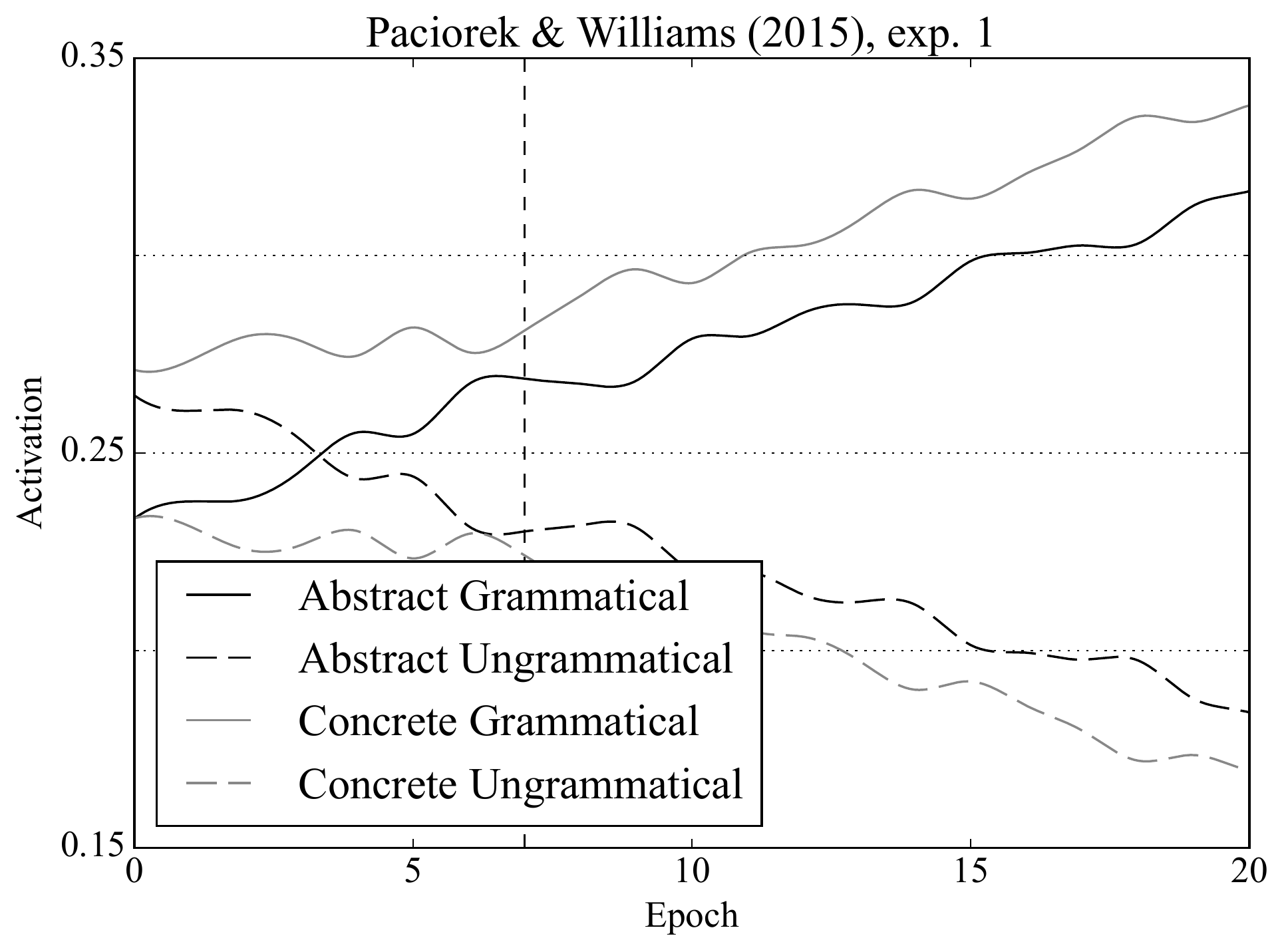}
			\end{minipage}
			\begin{minipage}{\columnwidth}
				\includegraphics[width=3.0in]{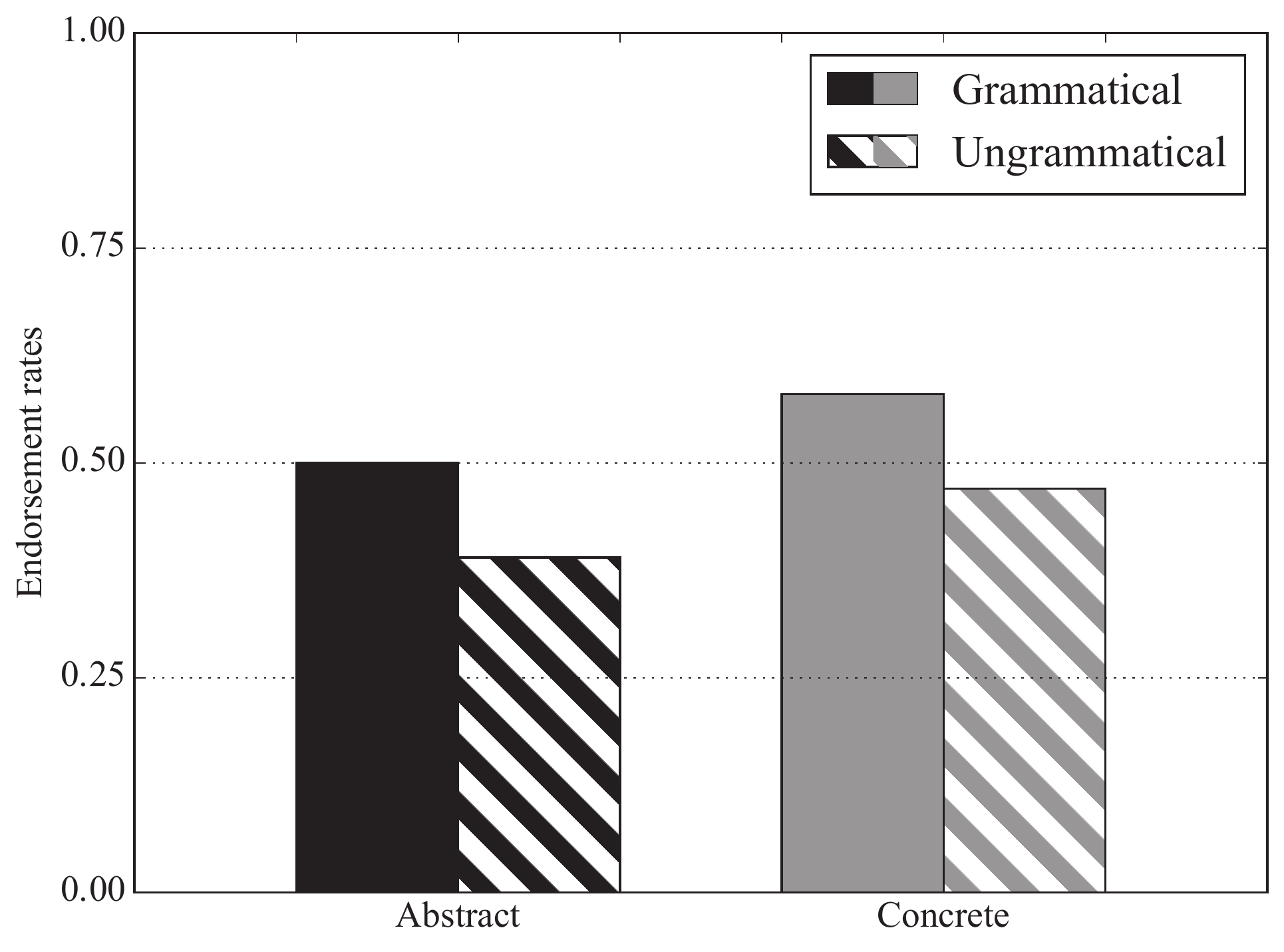}
			\end{minipage}
			\caption{Results of our simulation along with the behavioural results of Paciorek and Williams (2015), exp. 1. The hyperparameters used were the same as in the simulation of Williams (2005).}
\end{figure}

We obtained semantic representations using the \texttt{skip-gram} architecture \citep{Mikolov:2013uz} provided by the \texttt{word2vec} package,\footnote{\url{https://code.google.com/p/word2vec/}} trained with hierarchical softmax on the British National Corpus or on a Chinese Wikipedia dump file of comparable size.
The parameters used were as follows: window size: B5A5, vector dimensionality: $300$, subsampling threshold: $t = e^{-3}$ only for the English corpus.

The \texttt{skip-gram} model encapsulates the idea of distributional semantics introduced above by learning which contexts are more probable for a given word.
Concretely, it uses a neural network architecture, where each word from a large corpus is presented in the input layer and its context (i.e. several words around it) in the output layer.
The goal of the network is to learn a configuration of weights such that when a word is presented in the input layer the nodes in the output that become more activated correspond to those words in the vocabulary, which had appeared more frequently as its context.

As argued above, the resulting representations will carry, by means of their distributional patterns, semantic information such as concreteness or animacy.
Consistent with the above hypotheses, we predict that given a set of words in the training phase, the degree to which one can generalise to novel nouns will depend on how much the relevant concepts are reflected in the former words.
If, for example, the words used during the training session do not encode animacy based on their co-occurrence statistics, albeit denoting animate nouns, then generalising to other animate nouns would be more difficult.

\begin{figure}[t]
	\centering
			\begin{minipage}{\columnwidth}
        \includegraphics[width=\columnwidth]{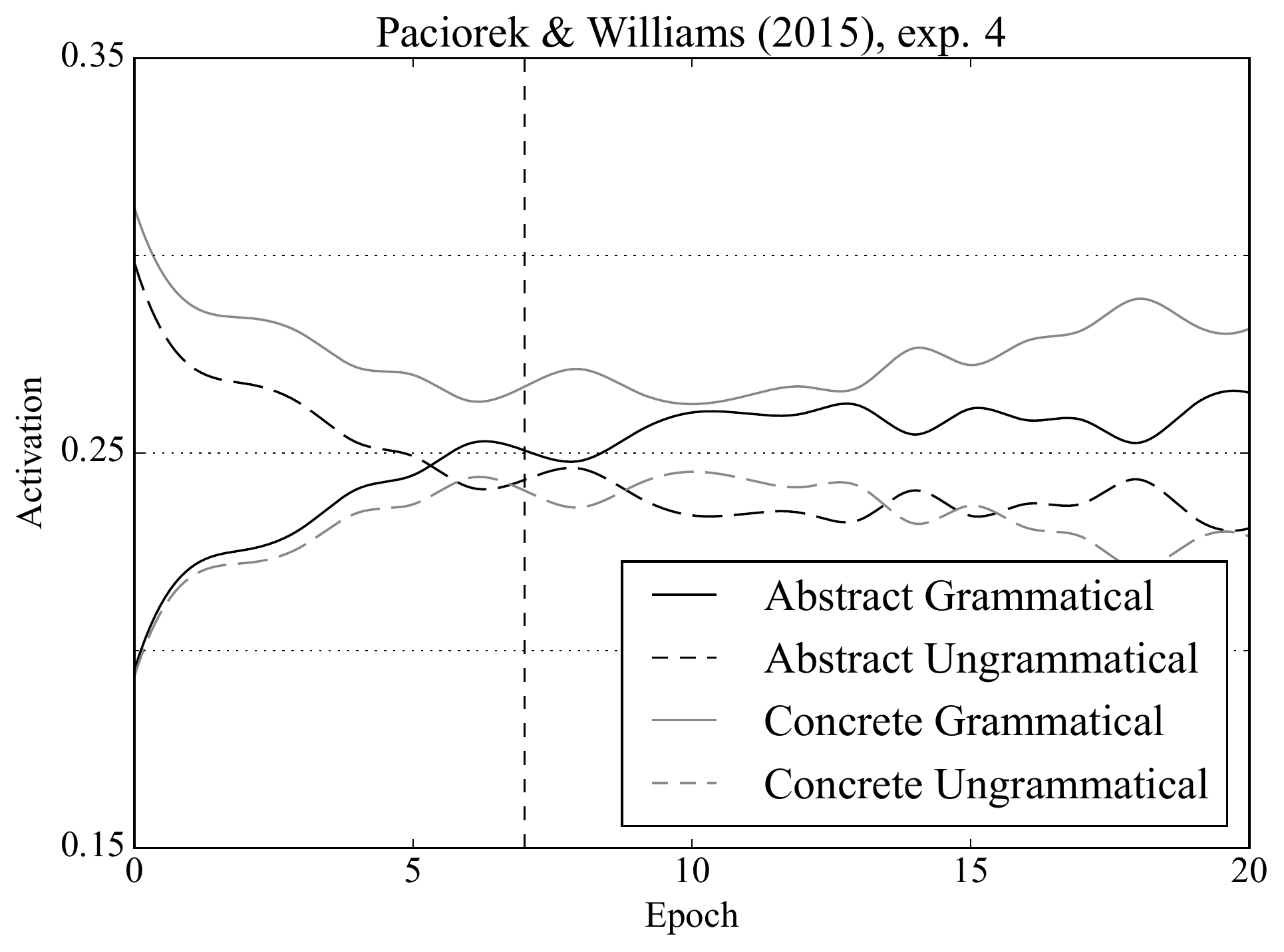}
			\end{minipage}
			\begin{minipage}{\columnwidth}
				\includegraphics[width=3.0in]{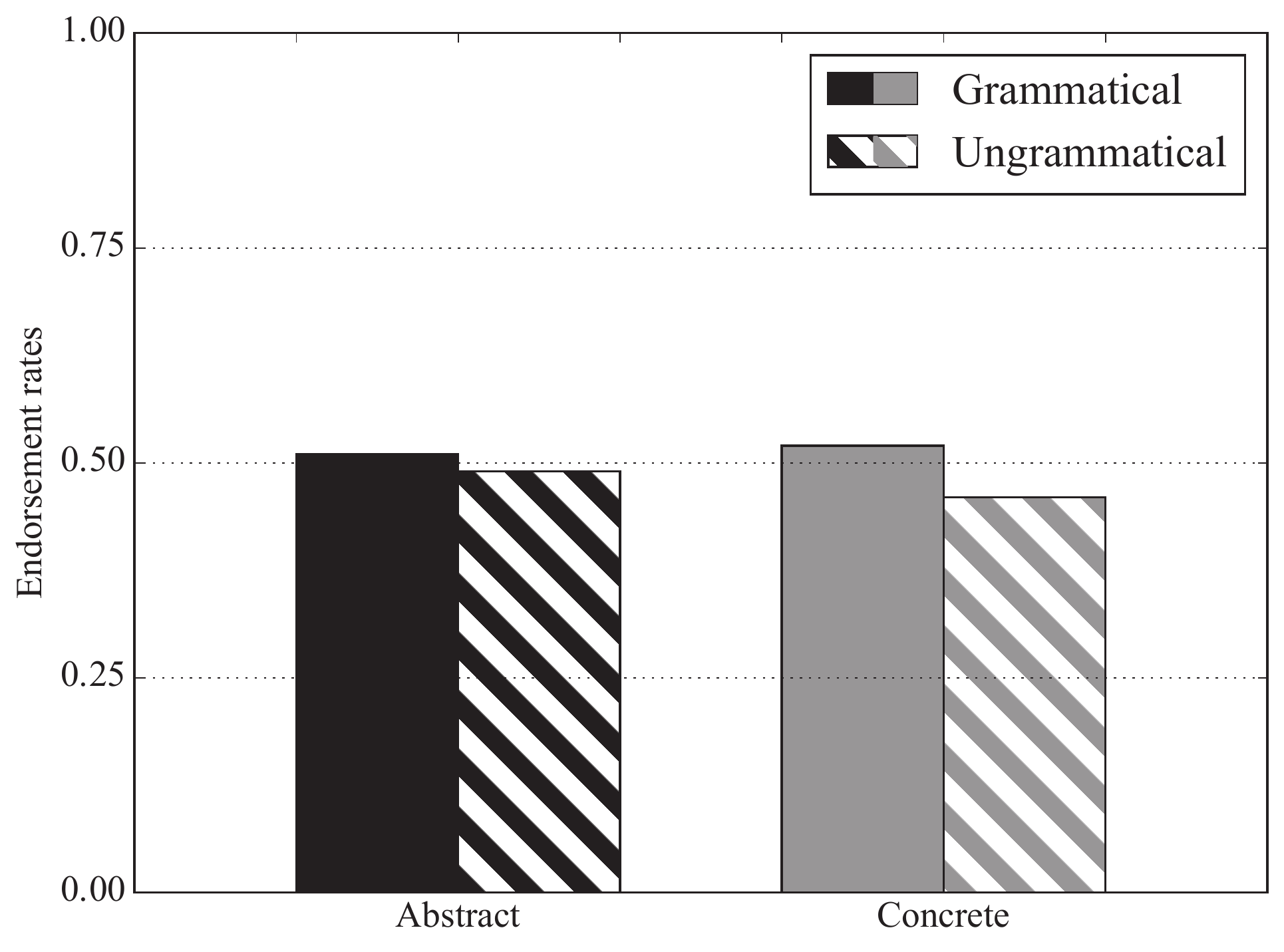}
			\end{minipage}
			\caption{Results of our simulation along with the behavioural results of Paciorek and Williams (2015), exp. 4. The hyperparameters used were the same as in the simulation of Williams (2005).}
\end{figure}

In order to examine this prediction, we fed the resulting semantic representations to a non-linear classifier (a feedforward neural network) the task of which was to learn to associate noun representations to determiners or verbs, depending on the study in question.
During the training phase, the neural network received as input the semantic vectors of the nouns and the corresponding determiners/verbs (coded as 1-in-$N$ binary vectors, where $N$ is the number of novel non-words)\footnote{All the studies reported use four novel non-words.} in the output vector.
Using backpropagation with stochastic gradient descent as the learning algorithm, the goal of the network was to learn to discriminate between grammatical and ungrammatical noun -- determiner/verb combinations.
We hypothesise that this could be possible if either specific features of the input representation or a combination of them contained the relevant concepts.
Considering the \emph{distributed} nature of our semantic representations, we explore the latter option by adding a $\tanh$ hidden layer, the purpose of which was to extract non-linear combinations of features of the input vector.
We then recorded the generalisation ability through time (epochs) of our classifier by simply asking what would be the probability of encountering a known determiner $\mathit{k}$ with a novel word $\vec{w}$ by taking the softmax function:

\begin{align}
	p(y=k|\vec{w}) = \frac{ \exp \left ( net_{k} \right ) } {\sum_{k\sp{\prime} \in K} \exp \left ( net_{k\sp{\prime}} \right )}.
	\label{eq:softmax_function}
\end{align}

\section{Results and Discussion}\label{sec:results_and_discussion}

Figures~1-4 show the results of the simulations across four different datasets which reflect different semantic manipulations.
The simulations show the generalisation gradients obtained by applying eq. (1) to every word in the generalisation set and then keeping track of the activation of the different determiners (W, L\&W) or verbs (P\&W) through time.
For example, in W where the semantic distinction was between animate and inanimate concepts `gi lion' would be considered a grammatical sequence while `ro lion' an ungrammatical one.

\begin{figure}[!t]
	\centering
			\begin{minipage}{\columnwidth}
				\includegraphics[width=\columnwidth]{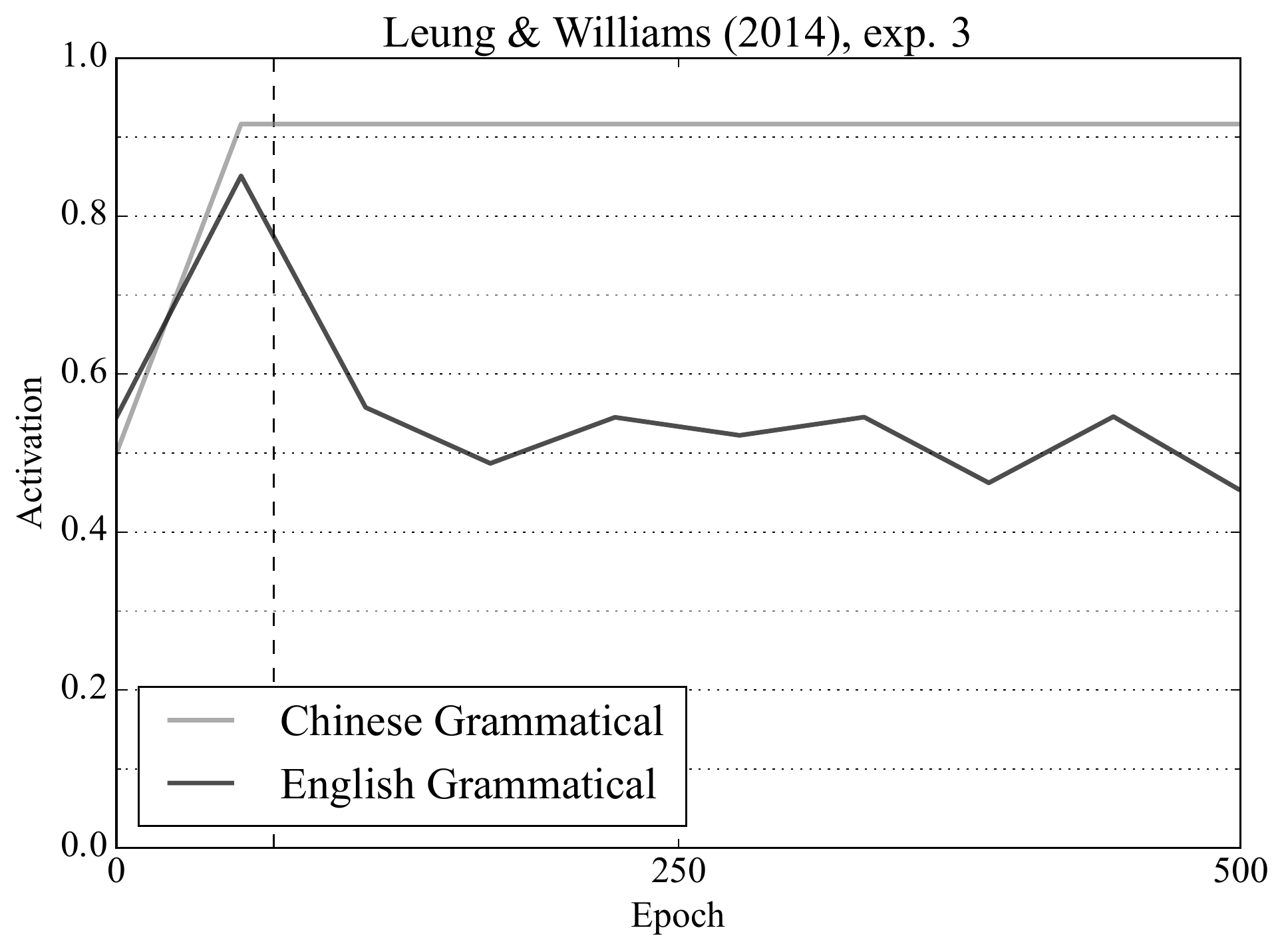}
			\end{minipage}
			\begin{minipage}{\columnwidth}
				\includegraphics[width=\columnwidth]{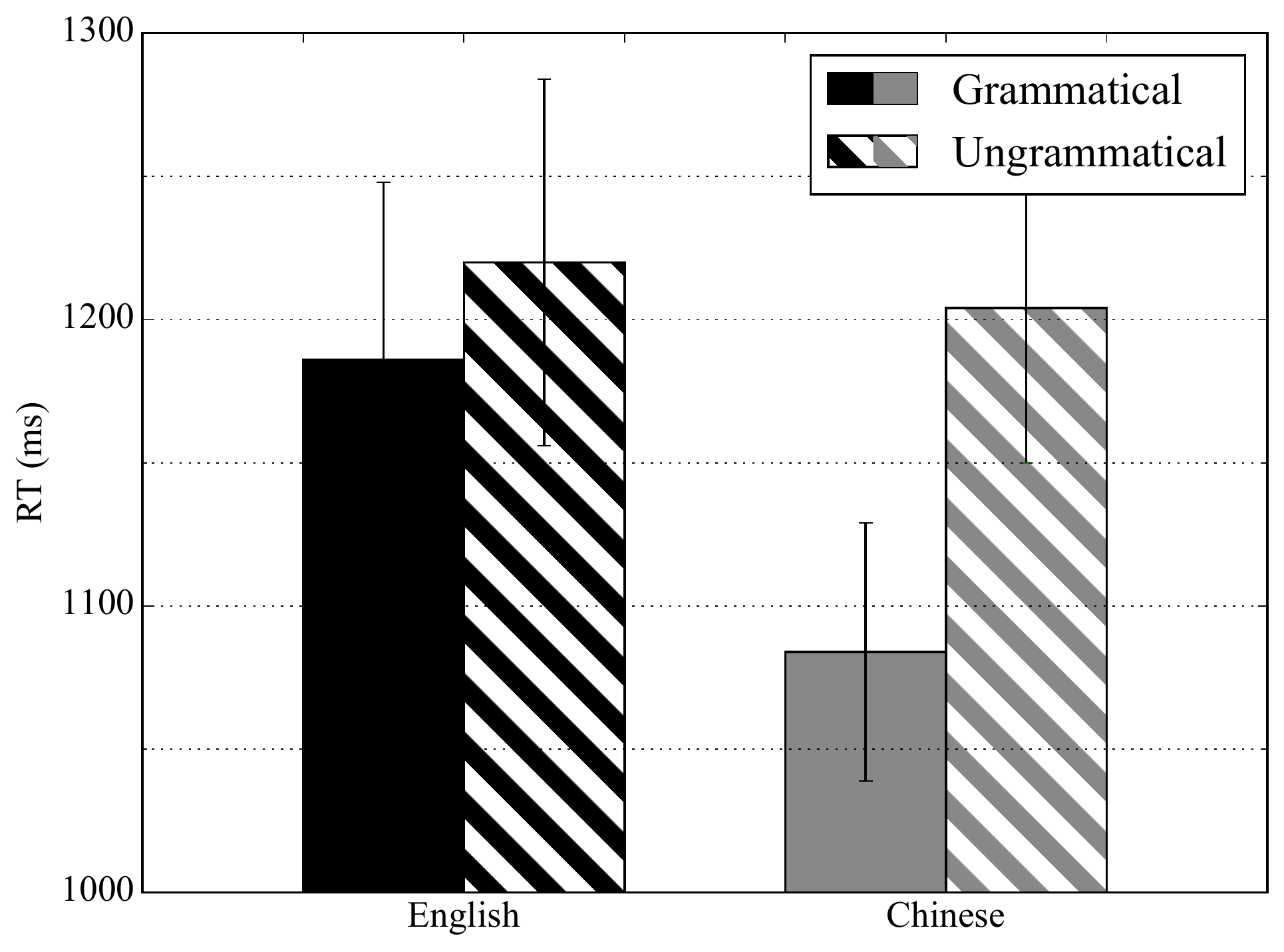}
			\end{minipage}
			\caption{Results from Leung and Williams (2014), exp. 3. See text for more info on the measures used. The gradients for the ungrammatical combinations are ($1 - \text{grammatical}$). The value of the weight decay was set to $\gamma = 0.05$ while the rest of the hyperparameters used were the same as in the simulation of Williams (2005).}
\end{figure}

If the model has been successful in learning that `gi' should be activated more given animate concepts then the probability $P(y=\text{gi}|\vec{w}_{\text{lion}})$ would be higher than $P(y=\text{ro}|\vec{w}_{\text{lion}})$.
Fig.~1 shows the performance of the classifier on the testing set of W where, in the behavioural data, selection of the grammatical item was significantly above chance in a two alternative forced choice task for the unaware group.
The slopes of the gradients clearly show that on such a task the model would favour grammatical combinations as well.

Figures~2-3 plot the results of two experiments from P\&W which focused on the abstract/concrete distinction.
P\&W used a \emph{false memory} task in the generalisation phase, measuring learning by comparing the endorsement rates between novel grammatical and novel ungrammatical verb-noun pairs.
It was reasoned that if the participants had some knowledge of the system they would endorse more novel grammatical sequences.
Expt~1 (Fig.~2) used generalisation items that were higher in semantic similarity to trained items than was the case in Expt~4 (Fig.~3).
The behavioural results from the unaware groups (bottom rows) show that this manipulation resulted in larger grammaticality effects on familiarity judgements in Expt~1 than Expt~4, and also higher endorsements for concrete items in general in Expt~1.
Our simulation was able to capture both of these effects.

L\&W Expt~3 examined the learnability of a system based on a long/flat distinction, which is reflected in the distributional patterns of Chinese but not of English.
In Chinese, nouns denoting long objects have to be preceded by a specific classifier while flat object nouns by another.
L\&W's training phase consisted of showing to participants combinations of thin/flat objects with novel determiners, asking them to judge whether the noun was thin or flat.
After a period of exposure, participants were introduced to novel determiner -- noun combinations, which either followed the grammatical system (\emph{control} trials) or did not (\emph{violation} trials).
Participants had significantly lower reaction times (Fig.~4, bottom row) when presented with a novel grammatical sequence than an ungrammatical sequence, an effect not observed in the RTs of the English participants.
The corresponding results of our simulations plotted in Fig.~4 show that indeed the regularity was learnable when the semantic model had only experienced a Chinese text, but not when it experienced the English corpus.

While more direct evidence is needed to support our initial hypothesis, our results seem to point to the direction that semantic information encoded by the distributional characteristics of words when found in large corpora can be important in determining what could be implicitly learnable.

\section*{Acknowledgments}

The first author is supported by the Onassis Foundation. We would like to
thank the three anonymous reviewers for their valuable feedback.

\bibliographystyle{apa}
\bibliography{bib}

\end{document}